\documentclass[conference]{IEEEtran}
\IEEEoverridecommandlockouts

\usepackage{cite}
\usepackage{amsmath,amssymb,amsfonts}
\usepackage{algorithmic}
\usepackage{graphicx}
\usepackage{textcomp}
\usepackage{xcolor}
\usepackage[colorlinks=true, linkcolor=black, citecolor=black, urlcolor=black]{hyperref}
\def\BibTeX{{\rm B\kern-.05em{\sc i\kern-.025em b}\kern-.08em
		T\kern-.1667em\lower.7ex\hbox{E}\kern-.125emX}}

\usepackage{booktabs}
\graphicspath{{./Images/}}
\usepackage{svg}
\usepackage{mathtools}

\begin{document}
	
	\author{
		\IEEEauthorblockN{Stelios Zarifis}
		\IEEEauthorblockA{\textit{Department of Electrical and Computer Engineering} \\
			\textit{National Technical University of Athens} \\
			Athens, Greece \\
			stelios.zarifis@gmail.com}
	}
	
	\title{Diffusion-Based Forecasting for Uncertainty-Aware Model Predictive Control \\
	}
	

	\author{Stelios Zarifis\textsuperscript{1,2}, Ioannis Kordonis\textsuperscript{2}, and Petros Maragos\textsuperscript{1,2}\\[0.5em]
		\textsuperscript{1}Robotics Institute, Athena Research Center, Athens, Greece\\
		\textsuperscript{2}School of Electrical and Computer Engineering, National Technical University of Athens, Athens, Greece\\[0.5em] \texttt{s.zarifis@athenarc.gr, kordonis@central.ntua.gr, petros.maragos@athenarc.gr}
	}
	
	\maketitle
	\begin{abstract}
		We propose \emph{Diffusion-Informed Model Predictive Control} (\emph{D-I MPC}), a generic framework for uncertainty‐aware prediction and decision-making in partially observable stochastic systems by integrating diffusion‐based time series forecasting models in Model Predictive Control algorithms. In our approach, a diffusion‐based time series forecasting model is used to probabilistically estimate the evolution of the system’s stochastic components. These forecasts are then incorporated into MPC algorithms to estimate future trajectories and optimize action selection under the uncertainty of the future. We evaluate the framework on the task of energy arbitrage, where a Battery Energy Storage System participates in the day‐ahead electricity market of the New York state. Experimental results indicate that our model‐based approach with a diffusion-based forecaster significantly outperforms both implementations with classical forecasting methods and model‐free reinforcement learning baselines.
	\end{abstract}
	
	
	\section{Introduction}
	Real-world decision‐making problems (for instance in energy markets, financial systems, inventory management) are characterized by high uncertainty and partial observability \cite{b1}. In such environments, the true state that actually influences the evolution of the system is unobservable. Hence, decision-makers rely on partial observations of the system. In this study, we highlight the potential of planning the control actions using a state-of-the-art probabilistic forecaster, over using traditional models or learning to act model-free, in a class of environments.
	
	We propose a framework that uses a diffusion‐based time series forecasting method to capture complex patterns in the stochastic dynamics, through probabilistic predictions. Integrating these probabilistic forecasts into a Model Predictive Control scheme, our algorithm aims to solve decision problems particularly in scenarios where data is sparse and there is no prior knowledge regarding how actions are taken. As a case study, we apply our method, to control a Battery Energy Storage System (BESS) in the day‐ahead electricity market of the New York state, where price volatility creates significant profit opportunities. We found that our method outperformed all other methods we tested. This work offers a flexible framework for uncertainty‐aware control using diffusion-based predictive models, applicable to various problems.
	
	\section{Background and Related Work}
	\subsection{Diffusion Models}
	\subsubsection{Denoising Diffusion Probabilistic Models}
	Denoising Diffusion Probabilistic Models (DDPMs) \cite{b2} are generative models that approximate a data distribution by iteratively denoising samples from a Gaussian prior. DDPMs are based on a forward process that gradually adds noise to the data and a learned reverse process that removes this noise step by step, aiming to reconstruct the initial data samples. These models became popular by achieving state-of-the-art results in modeling complex multimodal distributions in image generation tasks, thanks to their inherent high complexity and flexibility.
	
	The fundamental idea of DDPMs is to define a diffusion process that maps a data distribution to a simple prior distribution, usually a Gaussian, through a sequence of transformations. Given an initial data sample \( \mathbf{x}_0 \sim q(\mathbf{x}_0) \), the forward diffusion process iteratively adds Gaussian noise to acquire increasingly noisier versions of the sample:
	\vspace{-0.3em}
	\begin{equation}
		q(\mathbf{x}_t | \mathbf{x}_{t-1}) = \mathcal{N}(\mathbf{x}_t; \sqrt{\alpha_t} \mathbf{x}_{t-1}, (1 - \alpha_t) \mathbf{I}),
	\end{equation}
	where \( \alpha_t \) is a variance schedule controlling the noise level at each step \( t \). This process transforms the data into a nearly Gaussian distribution as \( t \to T \). To enable reconstruction, the forward process must be approximately invertible, which requires sufficiently small time steps so that the distribution shift between consecutive states remains smooth.
	
	A reverse process aims to denoise \( \mathbf{x}_T \) back to \( \mathbf{x}_0 \) through a learned model \( p_\theta \):
	\vspace{-0.3em}
	\begin{equation}
		p_\theta(\mathbf{x}_{t-1} | \mathbf{x}_t) = \mathcal{N}(\mathbf{x}_{t-1}; \mu_\theta(\mathbf{x}_t, t), \Sigma_\theta(t)).
	\end{equation}
	
%
%
	\vspace{-0.3em}
	Formulating DDPMs as latent variable generative models, their goal is to approximate the true data distribution \( q(\mathbf{x}_0) \) through the learned process:
	\vspace{-0.3em}
	\begin{equation}
		p_{\theta}\left(\mathbf{x}_{0}\right):=\int p_{\theta}\left(\mathbf{x}_{0: T}\right) d \mathbf{x}_{1: T} \approx q(\mathbf{x}_0), \label{eq:DDPM_goal}
		\vspace{-0.3em}
	\end{equation}
	where \( \mathbf{x}_{1}, \ldots, \mathbf{x}_{T} \) are latent variables of the same dimensionality as the data \( \mathbf{x}_{0} \sim q\left(\mathbf{x}_{0}\right) \).

%
	
	\subsubsection{TimeGrad Model}
	TimeGrad \cite{b3} is an autoregressive probabilistic forecasting model that combines DDPMs with a Recurrent Neural Network to model time series distributions. TimeGrad can capture multimodal distributions and long-range dependencies in multivariate time series better than traditional forecasters. A multivariate time series is defined as \( x_{i,k} \in \mathbb{R} \) for \( i \in \{1, \dots, D\} \), where \( k \) is the time index and \( i \) is the time series index, and the multivariate vector at time \( k \) is denoted as \( \mathbf{x}_k \in \mathbb{R}^D \). The forecasting problem can be defined as predicting a window of future values of a multivariate time series \( \mathbf{x}_{k_0:N} \) based on the past values \( \mathbf{x}_{1:k_0-1} \) and possibly additional information \( \mathbf{c}_{1:N} \) to encode events that affect the dynamics. In our framework, we will not use covariates, though they can be included if desired.
	
	TimeGrad models the conditional distribution of future values given past observations:
	\vspace{-0.3em}
	\begin{equation}
		q \left( \mathbf{x}_{k_0:N} | \mathbf{x}_{1:k_0-1} \right) = \prod_{k=k_0}^N q \left( \mathbf{x}_k | \mathbf{x}_{1:k-1} \right),
	\end{equation}
	where \(\mathbf{x}_k\) is the multivariate vector of the time series at time \(k\). Each factor is approximated using a DDPM conditioned on historical data, encoded using an RNN:
	\vspace{-0.3em}
	\begin{equation}
		\mathbf{h}_{k} = \text{RNN}_\theta \left( \mathbf{x}_{k}, \mathbf{h}_{k-1} \right),
	\end{equation}
	where $\mathbf{h}_{k-1}$ is the hidden RNN state at time $k-1$. The model learns to approximate the future time series distribution:
	\vspace{-0.3em}
	\begin{equation}
		p_\theta \left( \mathbf{x}_k | \mathbf{h}_{k-1} \right) \approx q \left( \mathbf{x}_k | \mathbf{x}_{1:k-1} \right). \label{eq:individual_TimeGrad_goal}
	\end{equation}
	Here, $\theta$ includes both RNN and diffusion model parameters, which are optimized jointly. Therefore, TimeGrad's goal is to approximate the conditional distribution of the future values using the individual learned distributions of Equation \eqref{eq:individual_TimeGrad_goal}:
	\vspace{-0.3em}
	\begin{equation}
		p_\theta\left(\mathbf{x}_{k_0:N} | \mathbf{x}_{1:k_0-1} \right) \, \mathclap{=} \prod_{k=k_0}^N p_\theta\left(\mathbf{x}_k | \mathbf{h}_{k-1}\right) \; \mathclap{\approx} \;\; q \left(\mathbf{x}_{k_0:N} | \mathbf{x}_{1:k_0-1} \right). \label{eq:TimeGrad_goal}
	\end{equation}
	Observing that TimeGrad's goal in Equation \eqref{eq:TimeGrad_goal} is the exact same as DDPMs' goal in Equation \eqref{eq:DDPM_goal}, the authors utilized the diffusion processes to learn the target distribution.
	
%
%
%
%
	
	\subsection{Reinforcement Learning}
	
	Reinforcement Learning (RL) is a framework for learning control policies from interactions in the environment.
	
	\emph{Model-free RL} value-based methods such as Q-learning \cite{b4} and Deep Q-Networks \cite{b5} and policy-based methods like Proximal Policy Optimization \cite{b6} have demonstrated success in complex tasks. \emph{Model-based RL} methods use an approximate model of the environment to plan actions. These methods are motivated by scenarios with limited data, where model-free approaches often struggle to learn a robust policy due to insufficient interactions with the environment \cite{b7}.
	
	Several ideas have been recently proposed on related tasks, especially in robotics. Approaches such as \cite{b8,b9} use probabilistic models to improve robustness and efficiency of planning. Moreover, recent works such as \cite{b10,b11,b12,b13,b14} utilize diffusion models to generate state-action sequences that optimize a given goal and \cite{b15} estimates the states based on noisy observations, leading to more robust control. However, most of the proposed models are trained on a dataset of state-action pairs, which is not always available. Our method is based on a model of the system, generated by a diffusion-based forecaster, to directly optimize the action sequence, without any prior knowledge on how actions are chosen, which is suitable for environments with sparse data or where it is not possible to make multiple simulations to generate training data.
	
	\section{Methodology}
	\subsection{Problem Definition}
	We present a method for a class of partially observable systems, where the dynamics can be separated into a deterministic and a stochastic component, and the actions of an agent do not affect the stochastic dynamics. This class of problems is interesting as it is widespread in many real-world applications, such as in energy markets, finance, inventory management, and robotics, where the systems exhibit a mix of predictable and uncertain dynamics \cite{b16}. Our method models the system as a Partially Observable Markov Decision Process (POMDP). The POMDP is defined by the tuple
	\( (\mathcal{S}, \mathcal{A}, T, R, \Omega, O, \gamma) \), where
	\begin{itemize}
		\item \(\mathcal{S}\) is the set of all possible (hidden) states,
		\item \(\mathcal{A}\) is the set of actions available to the decision-maker,
		\item \(T: S \times A \to \Delta(S)\) is the state transition function, with \(T(s' \mid s, a)\) giving the probability of transitioning to state \(s'\) when action \(a\) is taken in state \(s\), where \( \Delta (\mathcal{S}) \) denotes the probability distribution over the set of states S.
		\item \(R: O \times A \to \mathbb{R}\) is the reward function, yielding the immediate reward \(R(o, a)\) when taking action \(a\) in the observation state \(o\).
		\item \( \Omega \) is the set of observations that the agent can receive.
		\item \(O: S \times A \to \Delta(\Omega)\), the observation function; \(O(o \mid s', a)\), is the probability of observing \(o \in \Omega\) given that the environment transitions to state \(s'\), with action \(a\).
		\item \(\gamma \in [0,1)\) is the discount factor.
	\end{itemize}
	
	At each time step \(k\), the environment is in some state \(s_k \in \mathcal{S}\) (which is not directly observed by the agent). The agent selects an action \(a_k \in \mathcal{A}\); as a result, the environment transitions to a new state \(s_{k+1}\) according to the probability \(T(s_{k+1} \mid s_k, a_k)\), and the agent receives an observation \(o_{k+1} \in \Omega\), where \(o_{k+1} \sim O(o_{k+1} \mid s_{k+1}, a_k)\) and an immediate reward \(r_k = R(o_k, a_k)\).
	
	The objective of the agent at time \( k_0 \), given the history of observations, is to find a policy \(\pi ( o_1, o_2, \cdots, o_{k_0}) \) that maximizes its expected cumulative discounted reward:
	\vspace{-0.3em}
	\begin{equation}
		\underset{\pi}{\text{maximize }} \; \mathbb{E}\left[\sum_{k=k_0}^{\infty} \gamma^k r_k\right].
		\vspace{-0.3em}
	\end{equation}	
	
	In many practical applications (such as energy bidding, battery storage control, or inventory management) the state and dynamics are unmeasurable or unknown. For instance, in energy bidding, where participants submit bids to buy or sell electricity before the publication of the actual energy prices, it is impossible to obtain a true state, as the formation of future prices depends on factors like the decisions of the bidders, the energy demand, the weather, which are not fully predictable.
	
	The following analysis considers problems for which we assume that the true state \( s_k \in \mathcal{S} \) is not directly observed. Instead, for each time step \( k \), the agent receives an observation \( o_k \in \Omega \). Furthermore, the observations are partially deterministic and partially stochastic: \( o_k = \left( o_k^d, o_k^s \right) \), where the actions affect only the deterministic part \( o_{k+1}^d = f (o_k^d, a_k) \) (for example, in the application presented in Section \ref{sec:results}, the state-of-charge of a battery evolves deterministically while energy prices have stochastic dynamics). The agent forecasts the stochastic part of the future observations \( \hat{o}_{k+1}^s, \hat{o}_{k+2}^s, \dots \) to plan its actions.
	
	\subsection{Probabilistic Forecasting with Diffusion Models}
	To account for the inherent uncertainty and complexity that most real-world systems have, we employ TimeGrad to provide a stochastic model for the observations dynamics, by approximating the conditional probability distribution of future values given historical data:
	\vspace{-0.3em}
	\begin{equation}
		q(\mathbf{o}_{k_0:N}^s \mid \mathbf{o}_{1:k_0-1}^s) \approx p_\theta \left( \mathbf{o}_{k_0:N}^s \mid \mathbf{h}_{1:k_0-1} \right).
	\end{equation}

	\subsection{Model Predictive Control with Diffusion Models}
	We embed the forecasting diffusion model into a Model Predictive Control (MPC) framework to determine the optimal decision sequence in the POMDP, considering the forecasts as the stochastic model of the system. MPC is a control strategy that optimizes the evolution of dynamic systems by predicting future behavior and determining the optimal control actions over a finite time horizon. . In the deterministic MPC, TimeGrad is used as a point estimator by aggregating its probabilistic forecasts. Specifically, at each decision epoch, we define the forecast operator \( \mathcal{F} \) as:
	\vspace{-0.3em}
	\begin{equation}
		\hat{o}_{k+1}^s = \mathcal{F}(o_0^s, \dots, o_k^s) = \mathrm{median}\left\{ \hat{o}_{k+1}^{s, (i)} \right\}_{i=1}^M,
	\end{equation}
	where \(\{ \hat{o}_{k+1}^{s, (i)} \}_{i=1}^M\) is the set of \(M\) forecasted future observations generated by TimeGrad (the operator may alternatively use the mean). Then, the MPC optimization problem is formulated as:
	\vspace{-0.3em}
	\begin{equation}
		\begin{aligned}
		\underset{a_{k_0},\dots,a_{N + k_0 - 1}}{\text{maximize }} & \sum_{k=k_0}^{N + k_0 - 1} R(\hat{o}_k, a_k) \\
		\text{subject to} \quad & o_0, \dots, o_{k_0 - 1} \text{ observed}, \\
		& \hat{o}_{k+1}^d = f (\hat{o}_k^d, a_k), \; \hat{o}_{k_0} = o_{k_0}, \\
		& \hat{o}_{k+1}^s = \mathcal{F}(o_0^s, \dots, o_{k_0 - 1}^s, \hat{o}_{k_0}^s,\dots, \hat{o}_k^s), \\
		& a_k \in \mathcal{A}, \quad k=k_0,\dots,N + k_0 - 1.
	\end{aligned}
	\end{equation}
	
	By aggregating TimeGrad's probabilistic forecasts, our approach handles uncertainty only at the forecast level, resulting in a deterministic MPC optimization process. In contrast, stochastic MPC (SMPC) considers multiple trajectory realizations to integrate uncertainty in the optimization process too.
	
	\subsection{Stochastic Model Predictive Control with Diffusion Models}
	To handle stochasticity in the system dynamics, which impacts the decision sequence, we extend the deterministic MPC framework by considering multiple forecast trajectories throughout the optimization process. The distribution of generated future trajectories is used to model the uncertainty.
	
	We implement Monte Carlo SMPC, where we incorporate forecast uncertainty into the optimization process by sampling \( M \) trajectories from the future values distribution generated by TimeGrad. The goal is to maximize the expectation of the total reward, which is approximated by the average total reward of \( M \) realizations of the stochastic system. For each scenario \( i \in \{1, \dots, M\} \), the entire future trajectory from time \( k_0 \) to \( N + k_0 - 1 \) is generated recursively:
	\vspace{-0.3em}
	\begin{equation}
		\hat{o}_{k+1}^{s, (i)} \sim p_\theta \left( \hat{o}_{k+1}^{s, (i)} | \mathbf{h}_{k}^{(i)} \right), \quad k = k_0,\dots, N + k_0 - 1,
		\vspace{-0.3em}
	\end{equation}
	where \( p_\theta \) denotes TimeGrad's forecast distribution. The cumulative reward for each scenario is given by:
	\vspace{-0.3em}
	\begin{equation}
		\mathcal{J}^{(i)} = \sum_{k=k_0}^{N + k_0 -1} R\left(\hat{o}_k^{(i)}, a_k\right).
		\vspace{-0.3em}
	\end{equation}
	
	The SMPC optimization problem is then formulated as:
	\vspace{-0.3em}
	\begin{equation}
		\begin{aligned}
		\underset{a_{k_0},\dots,a_{N + k_0 - 1}}{\text{maximize }} & \quad  \frac{1}{M} \sum_{i=1}^{M} \mathcal{J}^{(i)} \\
		\text{subject to} \quad & o_0, \dots, o_{k_0 - 1} \text{ observed}, \\
		& \hat{o}_{k+1}^d = f (\hat{o}_k^d, a_k), \; \hat{o}_{k_0} = o_{k_0} \\
		& \hat{o}_{k+1}^{s, (i)} \sim p_\theta \left( \hat{o}_{k+1}^{s, (i)} | \mathbf{h}_{k}^{(i)} \right), \\
		& a_k \in \mathcal{A}, \quad k=k_0,\dots,N-1.
		\end{aligned}
		\vspace{-0.3em}
	\end{equation}
	This formulation models the uncertainty in both the forecasting and the optimization process, enabling the controller to plan more robust actions, especially when the dynamics are strongly non-linear.
	
	\section{Results}
	\label{sec:results}
	
	\subsection{Experimental Setup}
	We evaluate our method in the context of energy arbitrage for a region in New York State. The task is to control a Battery Energy Storage System (BESS), to make long-term profitable transactions. The environment is modeled as a POMDP. The observation \(o_k = \left( SoC_k, p_k \right) \in O\) consists of the battery's state-of-charge (SoC), and the electricity prices at time \( k \). The action \(a_k \in \mathcal{A}\) represents the amount of energy flowing in or out of the BESS, as a percentage of battery capacity. The transition function \( T \) has a deterministic and a stochastic part. The battery's SoC evolves deterministically, while the prices evolve stochastically. The reward is defined as the revenue generated from energy transactions minus the cost of battery degradation, according to a battery model, inspired by \cite{b17}. The action space is defined by the restrictions of the battery, i.e. the maximum energy flow, the battery's efficiency and capacity. We use hourly electricity prices from the New York Independent System Operator to test our method. The components of the POMDP are defined as:
	\vspace{-0.5em}
	\begin{equation}
		\begin{cases}
			s_k \in \mathcal{S}: \text{ unknown hidden state} \\
			a_k \in \mathcal{A}(o_k) = \left[ -\min \left( a_{max}, \frac{\text{SoC}k}{\eta} \right), \min \left( a_{max}, \frac{1 - \text{SoC}_k}{\eta} \right) \right] \\
			\begin{aligned}
				\hat{o}_{k+1} =& \left( SoC_{k+1}, \hat{o}_{k+1}^s \right) = \left( SoC_{k+1}, \hat{p}_{k+1}^s \right), \\
				& \text{SoC}_{k+1} = \text{SoC}_k + \eta \cdot a_k, \; \hat{o}_{k+1}^s = \mathcal{F}(\hat{o}_0^s, \dots, \hat{o}_k^s)
			\end{aligned} \\
			R \left( o_k, a_k \right) = 
			\text{Revenue}(o_k, a_k) - \text{Degradation}(o_k, a_k)
		\end{cases}
		\vspace{-0.3em}
	\end{equation}
	Optimal operation of grid‐connected battery energy storage systems over their lifetime has been also examined in~\cite{b18}.
	
	\subsection{Model Comparison and Observations}
	
\begin{figure}[h]
		\centering
		\vspace{-1.2em}
		\includesvg[width=\linewidth]{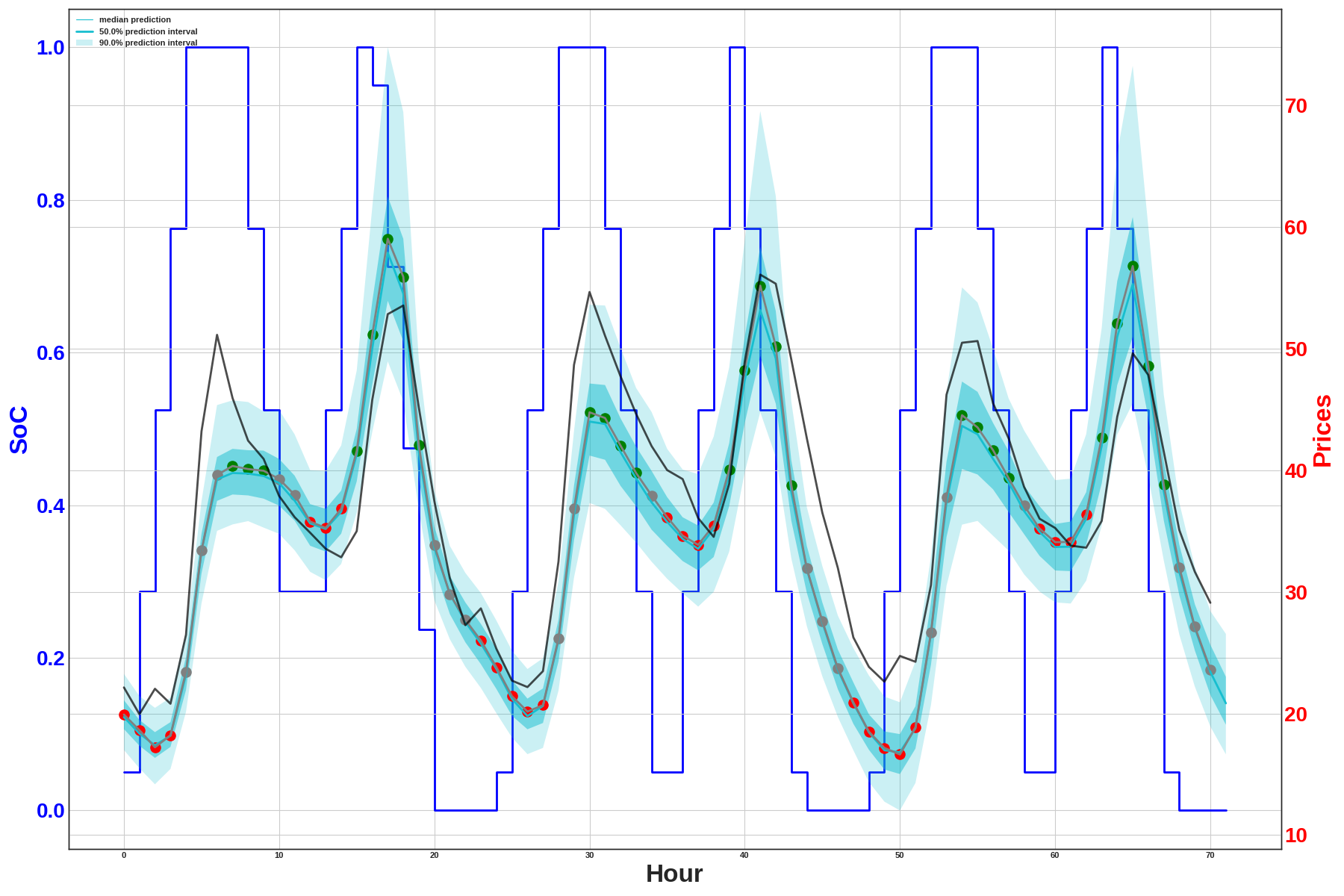}
		\vspace{-2.2em}
		\caption{\textit{Strategy planned by the MPC Optimizer. The dotted line represents the forecasted prices, the black line represents the actual electricity prices, the blue line represents the battery's SoC, and the cyan shaded areas represent the 90\% and 50\% quantiles of forecast distributions.}}
		\label{fig:mpc_example}
	\end{figure}

\begin{figure}[h]
		\centering
		\vspace{-2.2em}
		\includesvg[width=\linewidth]{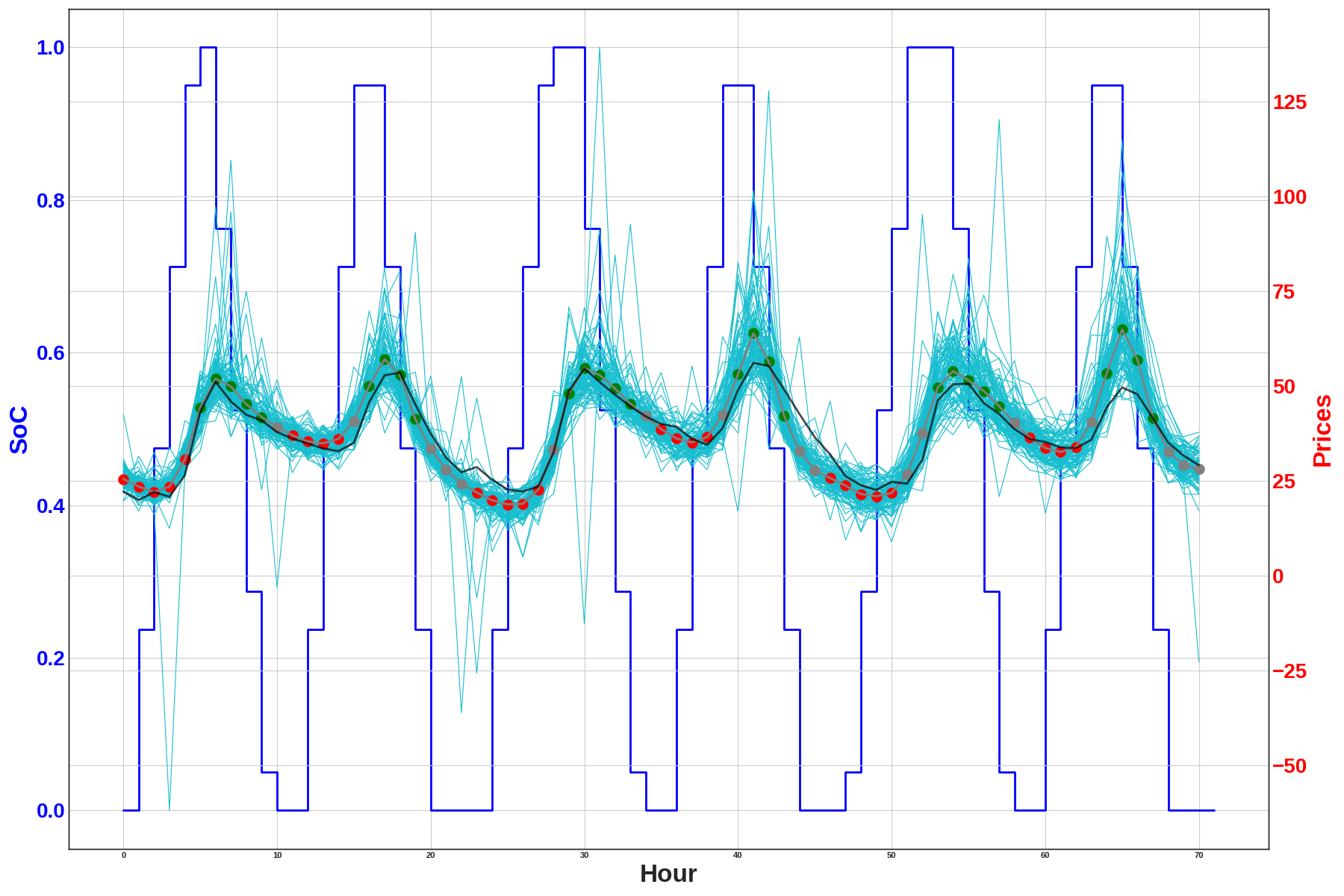}
		\vspace{-2.2em}
		\caption{\textit{Strategy planned by the SMPC Optimizer. The cyan lines represent the generated trajectories, the black line represents the actual prices (published in the future) and the blue line represents the battery's SoC.}}
		\label{fig:MC_smpc_example}
		\vspace{-1em}
	\end{figure}

	Table~\ref{tab:mpc} summarizes the performance of our MPC methods, Table~\ref{tab:classic_mpc} compares our methods with MPC guided by more classical predictive models, and Table~\ref{tab:rl_comparison} compares our method against a set of model-free RL baselines. Figures~\ref{fig:mpc_example}--\ref{fig:MC_smpc_example} illustrate the control strategies, where MPC and Monte Carlo SMPC (100 realizations) are used to plan over a 3-day look-ahead horizon. The optimizer re-plans its actions daily as new prices are published, applying only the first day's actions.
	
	\begin{table*}[t]
		\centering
		\begin{minipage}{0.34\textwidth}
			\centering
			\caption{D-I MPC Performance (higher is better).}
			\vspace{-1.2em}
			\label{tab:mpc}
			\resizebox{\textwidth}{!}{%
				\begin{tabular}{lcc|cc}
					\toprule
					\textbf{Month} & \textbf{Perfect} & \textbf{Oracle} & & \textbf{MC} \\
					& \textbf{MPC} & \textbf{MPC} & \textbf{MPC} & \textbf{SMPC} \\
					\midrule
					2018-06 & $99.98$ & $97.13$ & $90.72$ & $92.11$ \\
					2018-07 & $147.46$ & $141.56$ & $134.61$ & $138.75$ \\
					2018-08 & $123.11$ & $115.31$ & $111.62$ & $112.43$ \\
					2018-09 & $108.44$ & $102.32$ & $100.65$ & $100.53$ \\
					2018-10 & $104.05$ & $95.36$ & $103.48$ & $102.84$ \\
					2019-04 & $60.35$ & $54.62$ & $53.42$ & $53.51$ \\
					2020-12 & $68.85$ & $59.31$ & $60.58$ & $61.61$ \\
					2021-01 & $64.03$ & $59.00$ & $57.09$ & $56.60$ \\
					2021-02 & $109.34$ & $108.33$ & $95.46$ & $96.69$ \\
					2021-03 & $56.94$ & $56.35$ & $50.15$ & $54.54$ \\
					\midrule
					\textbf{Sum} & $942.57$ & $889.30$ & $749.90$ & $869.62$ \\
					\textbf{Average} & $94.26$ & $88.93$ & $74.99$ & $86.96$ \\
					\bottomrule
				\end{tabular}%
			}
		\end{minipage}
		\hfill
		\begin{minipage}{0.595\textwidth}
			\centering
			\caption{D-I MPC vs. Classical Models MPC Performance (higher is better).}
			\vspace{-1.2em}
			\label{tab:classic_mpc}
			\resizebox{\textwidth}{!}{%
				\begin{tabular}{lccccccc}
					\toprule
					\textbf{Month} & \textbf{TimeGrad} & \textbf{AR} & \textbf{ARIMA} & \textbf{SARIMA} & \textbf{VAR} & \textbf{CNN} & \textbf{LSTM} \\
					& \textbf{MPC} & \textbf{MPC} & \textbf{MPC} & \textbf{MPC} & \textbf{MPC} & \textbf{MPC} & \textbf{MPC} \\
					\midrule
					2018-06 & $\mathbf{90.72}$ & $87.59$ & $-1.28$ & $80.25$ & $65.84$ & $60.47$ & $74.84$ \\
					2018-07 & $\mathbf{134.61}$ & $109.56$ & $-3.11$ & $106.86$ & $68.81$ & $3.59$ & $38.13$ \\
					2018-08 & $\mathbf{111.62}$ & $93.55$ & $-2.55$ & $93.4$ & $61.02$ & $-15.46$ & $3.59$ \\
					2018-09 & $\mathbf{100.65}$ & $81.44$ & $1.33$ & $83.38$ & $60.77$ & $-1.19$ & $32.56$ \\
					2018-10 & $\mathbf{103.48}$ & $57.71$ & $-1.31$ & $62.07$ & $60.85$ & $38.76$ & $79.75$ \\
					2019-04 & $\mathbf{53.42}$ & $16.45$ & $-1.64$ & $25.34$ & $30.77$ & $18.73$ & $50.13$ \\
					2020-12 & $\mathbf{60.58}$ & $47.05$ & $-0.79$ & $35.27$ & $30.65$ & $12.54$ & $54.41$ \\
					2021-01 & $\mathbf{57.09}$ & $47.51$ & $-0.16$ & $38.39$ & $37.63$ & $-2.23$ & $42.87$ \\
					2021-02 & $\mathbf{95.46}$ & $56.69$ & $-4.39$ & $36.93$ & $56.99$ & $-19.42$ & $-46.61$ \\
					2021-03 & $50.15$ & $20.44$ & $-0.92$ & $22.24$ & $28.2$ & $-5.31$ & $\mathbf{50.88}$ \\
					\midrule
					\textbf{Sum} & $\mathbf{857.78}$ & $618.00$ & $-14.82$ & $584.14$ & $501.54$ & $90.48$ & $380.54$ \\
					\textbf{Average} & $\mathbf{85.78}$ & $61.80$ & $-1.48$ & $58.41$ & $50.15$ & $9.05$ & $38.05$ \\
					\bottomrule
				\end{tabular}%
			}
		\end{minipage}
	\end{table*}
	
	\begin{table*}[!htbp]
		\vspace{-1.2em}
		\centering
		\caption{Model-Free RL and Diffusion-Informed MPC Performances (higher is better).}
		\vspace{-1.2em}
		\label{tab:rl_comparison}
		\resizebox{0.82\textwidth}{!}{%
		\begin{tabular}{lccccc|c}
			\toprule
			& \multicolumn{5}{c|}{\textbf{Model-Free RL (Knowledge: 12 real prices ahead)}} & \textbf{Model-Based RL} \\
			\textbf{Month} & \textbf{DQN} & \textbf{DQN} & \textbf{DQN} & \textbf{DQN} & \textbf{DQN} & \textbf{TimeGrad} \\
			& {\tiny $32 \times 64 \times 32$} & {\tiny $64 \times 128 \times 64$} & {\tiny $128 \times 256 \times 128$} & {\tiny $64 \times 128 \times 128 \times 64$} & {\tiny $64 \times 128 \times 256 \times 128 \times 64$} & \\
			\midrule
			2018-06             & $8.40$    & $30.89$   & $21.34$   & $4.62$    & $61.28$   & $\textbf{92.11}$  \\
			2018-07             & $14.99$   & $53.44$   & $30.08$   & $8.80$    & $45.05$   & $\textbf{138.75}$ \\
			2018-08             & $6.14$    & $46.00$   & $10.64$   & $3.56$    & $25.57$   & $\textbf{112.43}$ \\
			2018-09             & $3.87$    & $28.21$   & $17.99$   & $3.07$    & $20.14$   & $\textbf{100.53}$ \\
			2018-10             & $8.29$    & $31.75$   & $18.96$   & $8.79$    & $14.46$   & $\textbf{102.84}$ \\
			2019-04             & $3.43$    & $18.65$   & $10.25$   & $2.41$    & $6.00$    & $\textbf{53.51}$  \\
			2020-12             & $3.55$    & $15.35$   & $9.32$    & $3.67$    & $7.18$    & $\textbf{61.61}$  \\
			2021-01             & $2.91$    & $19.37$   & $10.47$   & $7.01$    & $10.62$   & $\textbf{56.60}$  \\
			2021-02             & $2.58$    & $12.44$   & $14.39$   & $3.36$    & $0.32$    & $\textbf{96.69}$  \\
			2021-03             & $1.59$    & $13.10$   & $7.46$    & $6.43$    & $5.65$    & $\textbf{54.54}$  \\
			\midrule
			\textbf{Sum}        & $55.76$   & $269.15$  & $151.80$  & $51.02$   & $196.27$  & $\textbf{869.62}$ \\
			\textbf{Average}    & $5.58$    & $26.92$   & $15.18$   & $5.10$    & $19.63$   & $\textbf{86.96}$  \\
			\bottomrule
		\end{tabular}
		}
		\vspace{-1.7em}
	\end{table*}
	
	\paragraph{Visualization of the Method}
	Figures~\ref{fig:mpc_example}--\ref{fig:MC_smpc_example} show that the SoC rises at low prices (red dots) and decreases at high prices (green dots), while no action is indicated by gray color.
	
	For both implementations, the algorithm adapts to daily market updates by planning over the horizon and applying only the first day's actions, since in the next day, new prices will be published, thus more accurate forecasts will be available.
	
	\paragraph{Diffusion-Informed MPC Performance}
	Table~\ref{tab:mpc} compares our implementations with two idealized ones. The Perfect MPC algorithm performs MPC on a full range of prices at once (has full knowledge of the future), while Oracle MPC uses as model the real prices (perfect forecaster) for the same horizon as the rest MPC algorithms. MPC and Monte Carlo MPC with a diffusion-based forecaster closely match the performance of the idealized benchmarks, having reward gaps of merely 3.54\% and 2.22\% with Oracle MPC (MPC with a perfect forecaster), respectively, confirming that diffusion models are ideal to guide the MPC optimization process.
	
	
	\paragraph{Classical Model MPC vs. Diffusion-Informed MPC}
	
	When comparing forecasting models in Table~\ref{tab:classic_mpc}, MPC guided by a diffusion-based forecaster significantly outperforms those based on classical methods, with a 38.8\% margin over the second best. Notably, the LSTM of Table~\ref{tab:classic_mpc} has the exact same architecture with TimeGrad's RNN, to showcase the advantage of using diffusion processes for more accurate forecasts.
	
\begin{figure}[h]
		\centering
		\vspace{-1.3em}
		\includesvg[width=0.84\linewidth]{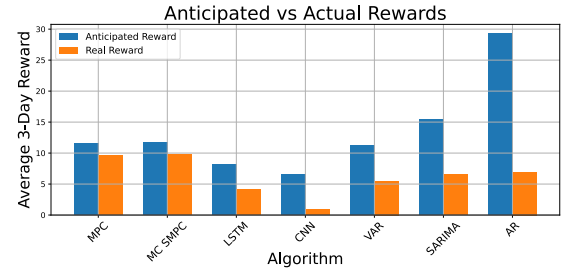}
		\vspace{-1.2em}
		\caption{Anticipated and actual rewards (3-day windows) for various models, sorted by how closely the anticipated matches the actual reward.}
		\vspace{-0.8em}
		\label{fig:expected_vs_real_reward}
	\end{figure}

	It is important to note that, although the deep learning models we trained achieve better error metrics compared to autoregressive models, their ability to guide MPC algorithms is significantly worse. The reason lies in the specific application on which we test our method, since accurately predicting the peaks and valleys of energy prices is far more critical than merely achieving low error rates. The CNN and LSTM models tended to estimate the trends of the price time series sometimes with a small shift, resulting in the agent, which follows the MPC policy, being confused on when it is optimal to make transactions. This is illustrated in Figure~\ref{fig:expected_vs_real_reward}, where the CNN and LSTM demonstrate better reward anticipation compared to classical models, since they provide more accurate forecasts. However, MPC based on these deep learning models does not exhibit improved performance, due to the said shifts. This underscores again the importance of accurately predicting the details in the state trajectory, hence diffusion-based forecasters are highly suitable to guide the MPC.
	
	
	\paragraph{Model-Free RL vs. Model-Based Approaches}
	Table \ref{tab:rl_comparison} shows that model-free RL implementations, even when provided with perfect knowledge of the next 12 prices as state, fall considerably behind \textit{Diffusion-Informed MPC}. The average reward of the best DQN agent is 69.5\% lower than the reward achieved by our model, highlighting the advantage of our model-based method over model-free approaches, since the system dynamics can be approximated by a model.
		
	
	\section{Conclusion}
	We defined a novel framework for uncertainty-aware prediction and decision‐making in partially observable stochastic systems by integrating diffusion‐based time series forecasting with Model Predictive Control. Our approach uses diffusion models to generate probabilistic forecasts of the system's stochastic dynamics, which are used to guide MPC algorithms to select optimal actions. The application of our method in a real-world scenario demonstrates significant advantages over MPC with traditional forecasters and model‐free RL approaches. Future work will explore further enhancements, including Scenario-Tree based MPC and a hybrid approach combining model-free and model-based methods.

\end{document}